\documentclass[runningheads]{llncs}
\usepackage{graphicx}
\usepackage{amsmath}
\usepackage{amssymb}
\usepackage[T1]{fontenc}
\usepackage{booktabs}
\usepackage{epsfig, multirow, tabularx, xcolor}
\usepackage{pifont}
\usepackage{hyperref}

\newcolumntype{C}{>{\centering\arraybackslash}p{2.5em}}

\usepackage{graphicx}
\begin{document}
\title{A Multi-Stream Fusion Network for Image Splicing Localization}
%
%

\author{Maria Siopi \and Giorgos Kordopatis-Zilos \and Polychronis Charitidis \and Ioannis Kompatsiaris \and Symeon Papadopoulos}
\authorrunning{M. Siopi et al.}
%
\institute{Information Technologies Institute, CERTH, Thessaloniki 60361, Greece
\email{\{siopi,georgekordopatis,charitidis,ikom,papadop\}@iti.gr}}
\maketitle              
\begin{abstract}
In this paper, we address the problem of image splicing localization with a multi-stream network architecture that processes the raw RGB image in parallel with other handcrafted forensic signals. Unlike previous methods that either use only the RGB images or stack several signals in a channel-wise manner, we propose an encoder-decoder architecture that consists of multiple encoder streams. Each stream is fed with either the tampered image or handcrafted signals and processes them separately to capture relevant information from each one independently. Finally, the extracted features from the multiple streams are fused in the bottleneck of the architecture and propagated to the decoder network that generates the output localization map. We experiment with two handcrafted algorithms, i.e., DCT and Splicebuster. Our proposed approach is benchmarked on three public forensics datasets, demonstrating competitive performance against several competing methods and achieving state-of-the-art results, e.g., 0.898 AUC on CASIA.

\keywords{image splicing localization \and image forensics \and multi-stream fusion network \and late fusion deep learning}
\end{abstract}

\section{Introduction}
\label{sec:intro}

Images have long been considered reliable evidence when corroborating facts. However, the latest advancements in the field of image editing and the wide availability of easy-to-use software create very big risks of image tampering by malicious actors. Moreover, the ability to easily alter the content and context of images especially in the context of social media applications further increases the potential use of images for disinformation. This is especially problematic as it has become almost impossible to distinguish between an authentic and tampered image by manual inspection. 

To address the problem, researchers have put a lot of effort on the development of image forensics techniques that can automatically verify the authenticity of multimedia. Nevertheless, capturing discriminative features of tampered regions with multiple forgery types (e.g., splicing, copy-move, removal) is still an open challenge \cite{korus2017digital}, especially in cases that the image in question is sourced from the Internet. Internet images have typically undergone many transformations (e.g. resizing, recompression), which result in the loss of precious forensics traces that could lead to the localization of tampered areas \cite{Zampoglou2017}. This work focuses on the localization of splicing forgeries in images, where a foreign object from a different image is inserted in an original untampered one.

Several approaches have been proposed in the literature, both \textit{handcrafted} and \textit{deep-learning}, that attempt to tackle the problem of image splicing localization. Handcrafted approaches \cite{Lin2009,Cozzolino2015} aim at detecting the manipulations by applying carefully-designed filters that highlight traces in the frequency domain or by capturing odd noise patterns in images. On the other hand, the more recent deep learning approaches \cite{Bappy2019,Wu2019,Hu2020} leverage the advancements in the field and build deep networks, usually adapting encoder-decoder architectures, trained to detect the tampered areas in images based on large collections of forged images. 

Although most splicing localization methods rely on handcrafted or deep learning schemes and work directly with the RGB images \cite{Lin2009,Cozzolino2015,Bappy2019,Mazaheri2019,Hu2020,ZhangY2021}, there are only few works that combine the two solutions using a single network to fuse the information extracted from the raw image and/or several handcrafted signals \cite{fontani2013framework,Wu2019,Charitidis2021}. The latter methods usually stack/concatenate all signals together in a multi-channel fashion and process them simultaneously by a single network stream that combines evidence from all signals to generate the output. This can be viewed as a kind of early fusion. Yet, some traces can be missed by the network when all signals are processed together. Instead, a late fusion approach could be employed, where each handcrafted signal along with the raw image is fed to a different network stream and then fused within the network to derive the output localization map. Each input signal is processed independently, and the network is able to capture the relevant information with different streams focused on a specific signal.

Motivated by the above, in this paper, we propose an approach that leverages the information captured from extracted handcrafted signals and the tampered images themselves. We develop a multi-stream deep learning architecture for late fusion following the encoder-decoder scheme. More specifically, the RGB images along with several handcrafted forensic signals, i.e., DCT \cite{Lin2009} for a frequency-based and Splicebuster \cite{Cozzolino2015} for noise-based representation, which are robust to the localization of the tampered areas with splicing manipulation \cite{Charitidis2021}. All signals are fed to different encoder streams that generate feature maps for each input signal. All extracted feature maps are then concatenated and propagated to a decoder network that fuses the extracted information and generates an output pixel-level map, indicating the spliced areas. By leveraging separate network streams for each input, we are able to extract richer features and, therefore, have more informative representations for the localization.

Our contributions can be summarized in the following:
\begin{itemize}
    \item We address the splicing localization problem with a multi-stream fusion approach that combines handcrafted signals with the RGB images.
    \item We build an encoder-decoder architecture that processes each signal in a different encoder stream and fuses them during the decoding.
    \item We provide a comprehensive study on three public datasets, where the proposed approach achieves state-of-the-art performance. 
\end{itemize}

\section{Related Work}
\label{sec:relwork}

Image splicing localization has attracted the interest of many researchers in the last few decades; hence, several solutions have been proposed in the literature. The proposed methods can be roughly classified into two broad categories, i.e., handcrafted and deep learning. 

Early image manipulation detection methods were designed to tackle the splicing localization problem using handcrafted methods, consisting of a simple feature extraction algorithm and can be categorized according to the signals they use and the compression type of the images they are applied on. For example, there are noise-based methods that analyse the noise patterns within images, e.g., Splicebuster \cite{Cozzolino2015} and WAVELET \cite{mahdian2009using}, methods that work with raw images analyzing them using different JPEG compression parameters and detect artifact inconsistencies, e.g., GHOST \cite{farid2009exposing} and BLOCK \cite {li2009passive}, and double quantization-based algorithms operating in the frequency domain, e.g., Discrete Cosine Transform (DCT) \cite{Lin2009}. An extensive review of such methods can be found in \cite{Zampoglou2017}. Nevertheless, the forensic traces captured by these algorithms can be easily erased by simple resizing and re-encoding operations. Also, these methods are often outperformed by their deep learning-based counterparts.

Later works use deep learning to localize splicing forgeries based on a neural network, extracting features only from the raw images. A seminal work in the field is ManTra-Net \cite{Wu2019}, which consists of two parts, a feature extractor and an anomaly detection network. The feature extractor computes features of the image by combining constrained CNNs, SRM features, and classical convolutional features concatenating them in a multi-channel fashion so as to be processed by the rest of the network. The detection network applies deep learning operations (LSTMs and CNNs) to the features extracted and exports the final localization map. SPAN \cite{Hu2020} advanced ManTra-Net and modeled the spatial correlation between image regions via local self-attention and pyramid propagation. In \cite{Bappy2019}, the model utilizes both the information from the frequency and the spatial domain of images. A CNN extracts the features in the spatial domain, while a Long Short-Term Memory (LSTM) layer receives the resampled features extracted from the image patches as input. The outputs of the two streams are fused into a decoder network that generates the final localization map. Mazaheri \cite{Mazaheri2019} added a skip connection to the above architecture, which exploits low-level features of the CNN and combines them with high-level ones in the decoder. In \cite{ZenanShi2020}, the authors followed an encoder-decoder architecture and introduced a bidirectional LSTM layer and gram blocks. The method proposed in \cite{ZhangY2021} combines top-down detection methods with a bottom-up segmentation-based model. In \cite{hao2021}, the authors proposed a multi-scale network architecture based on Transformers, exploiting self-attention, positional embeddings, and a dense correction module. However, the above architectures utilize only the information that can be extracted from the raw images, which can be boosted with the use of handcrafted signals. Only ManTra-Net leverages some handcrafted features, which, however, are early fused with image features in a multi-channeled manner. 

Finally, there are fusion approaches that leverage several handcrafted forensic signals \cite{fontani2013framework,Iakovidou2020,Charitidis2021} aiming to increase the robustness of the models. In \cite{fontani2013framework}, the authors employed the Dempster-Shafer theory of evidence \cite{gordon1984dempster} that allows handling uncertain predictions provided by several image forensics algorithms. In \cite{Iakovidou2020}, the authors proposed a handcrafted approach that extracts several handcrafted signals that are further refined to generate the output map. In \cite{Charitidis2021}, the authors proposed a deep learning-based architecture based on an encoder-decoder scheme that receives several maps from handcrafted signals concatenated in a multi-channel way and processed by a single-stream network. The latter two works do not exploit the information from the raw images into the fusion process and do not rely on multi-stream processing of the signals.
\section{Approach overview}
\label{sec:model}

\begin{figure*}[t]
    \centering
    \includegraphics[width=\textwidth]{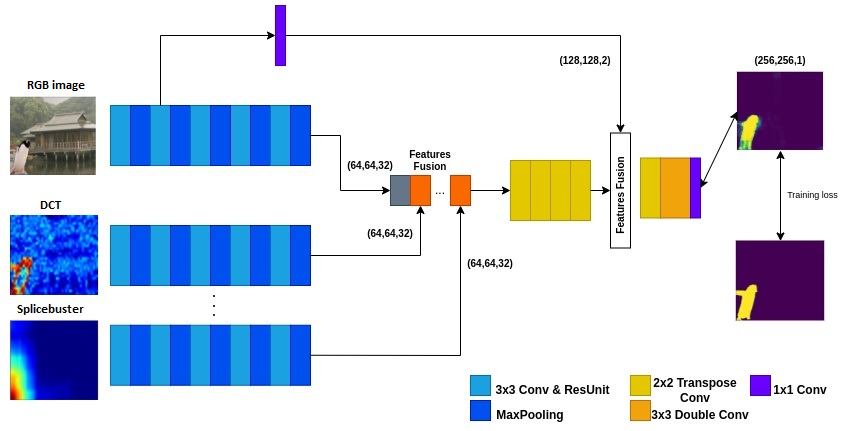}
    \caption{Overview of the proposed network architecture (best viewed in colour). The inputs are the RGB image and the handcrafted signals, i.e., DCT and Splicebuster, processed by a different encoder stream. The encoding outputs are fused and propagated to the decoder that outputs the predicted localization map, which is compared to the ground-truth mask to compute the loss.}
    \label{fig:architecture}
\end{figure*}

The main objective of our work is to develop a model that fuses different handcrafted forensic signals - in this work, we explore DCT \cite{Lin2009} and Splicebuster \cite{Cozzolino2015} - along with the raw manipulated image. The model follows an encoder-decoder architecture. In the decoder, we fuse the outputs of multiple encoding streams, which extract features of the input signals through convolutional operations. Figure \ref{fig:architecture} illustrates an overview of the proposed architecture.

\subsection{Multi-stream architecture}

Our architecture has two parts, i.e., an encoder and a decoder. For the encoder, we build multiple streams that process either the RGB images or the employed handcrafted signals. For each stream, we employ a network architecture similar to the one proposed in \cite{Mazaheri2019}. More specifically, each encoder stream comprises five stages consisting of a $3\times3$ convolutional layer, a residual block, and a max-pooling layer. The residual blocks are composed of two $3\times3$ convolutional layers with batch normalization \cite{ioffe2015batch} and a ReLU activation in the output. The number of channels of each stage output are [32, 64, 128, 256, 512]. At the end of each stream, we apply a $3\times3$ convolution with 32 output channels. Finally, we have two kinds of encoder streams, with and without the skip connection, as proposed in \cite{Mazaheri2019}. We use an encoder stream with the skip connection for the RGB image, while the remaining streams of the handcrafted signals do not have that skip connection. We empirically found that this setup yields the best performance.

For the decoder part, we first concatenate the feature maps of the encoder streams following a late-fusion approach, and we then process them by the main decoder network. The size of the decoder input depends on the number of encoder streams in the system. Unlike prior works \cite{Bappy2019,Mazaheri2019}, we build a more sophisticated architecture for our decoder, which performs upsampling in a learnable way by employing trainable transpose convolutional layers. We use as many transpose convolutional layers as the number of stages in our encoder streams, i.e., five layers with output channels [64, 32, 16, 2, 2]. Following the practice of \cite{Mazaheri2019}, 
We add a skip connection from the second stage of the image encoder stream and concatenate it to the feature map of the fourth decoder layer. At the end of the decoder, we apply two $3\times3$ convolutions with a number of channels equal to 2. The output aggregated pixel-level predictions derive from the application of a final $1\times1$ convolution with a single output channel, followed by a sigmoid activation that maps the values to the $[0, 1]$ range.

In that way, we build a multi-stream architecture that encodes several signals independently, i.e., RGB images and handcrafted signals, and performs a late-fusion in the model's bottleneck. The extracted features are then processed altogether by a decoder network that outputs a binary mask with per-pixel predictions without the need for further post-processing.

\subsection{Handcrafted signals}

In our approach, the number of the encoding streams equals the number of the handcrafted forensics signals used for the prediction, plus one for the manipulated image. Previous works tried to utilize the information from the frequency and the spatial domain of the processed images, combining schemes based on CNN and LSTM layers \cite{Bappy2019,Mazaheri2019}. In contrast, to capture information from the frequency domain, we employ the DCT \cite{Lin2009} handcrafted signal, which is a Fourier-based transform and can represent the JPEG images in the frequency domain. It divides the image into segments based on its resolution and applies the discrete cosine transform whose coefficients contain information related to the frequencies in the segments. DCT has been widely used in many different applications, including forgery detection. In that way, the input images are represented in the frequency domain, and with the application of convolutional filters, we capture spatial information from the frequency domain.

Other works, e.g., ManTra-Net \cite{Wu2019}, extract noise-based handcrafted signals, combined with the RGB image in a multi-channel way for image tampering localization. To this end, in this work, we employ the Splicebuster \cite{Cozzolino2015} as a noise-based handcrafted signal, which is among the top-performing handcrafted algorithms. Splicebuster extracts a feature map for the whole image in three steps: (i) compute the residuals of the image through a high-pass filter, (ii) quantize the output residuals, and (iii) finally generate a histogram of co-occurrences to derive a single feature map. Similar to DCT, we generate noise-based representations for the input images, and with the application of convolutional filters, we extract spatial information from these representations.

\subsection{Training process}

The signals are extracted using the publicly available service in \cite{Zampoglou2016}. During training, the RGB images and the extracted handcrafted signals are fed to the model, each in a different stream, and it outputs a binary map as a pixel-level prediction. The loss function used for the end-to-end training of the network is the binary cross-entropy loss computed based on the ground-truth masks and the generated outputs.
\section{Experimental setup}
\label{sec:exp}
This section describes the datasets used for training and evaluation of our models, the implementation details, and the evaluation metrics used in our experiments to measure the splicing localization performance.

\subsection{Datasets}
\label{sec:datasets}
For the training of our model, we use the Synthetic image manipulation dataset \cite{Bappy2019}, and we extract the maps of our handcrafted signals for each image in the dataset. The synthetic dataset contains images with tampered areas from splicing techniques.

For evaluation, we used three image manipulation datasets, i.e., CASIA \cite{casia}, IFS-TC \cite{ifs} and Columbia \cite{columbia}. The models are further fine-tuned to evaluation datasets. For CASIA, for the fine-tuning of our models we use CASIA2, which includes 5,123 tampered images, and for evaluation CASIA1, which includes 921 tampered images, i.e. we use only the subset with spliced images for evaluation. Regarding IFS-TC, we split the dataset to training and test set, and for fine-tuning we use the training set, which includes 264 tampered images, and for evaluation we use the test set, which includes 110 tampered images. Columbia \cite{columbia} is a very small dataset (180 tampered images); hence, we use it in its entirety for evaluation without fine-tuning our model.

\subsection{Implementation details}
\label{sec:imp_details}
All of the models have been implemented using  PyTorch~\cite{paszke2019}. For the training of our model, we use Adam \cite{kingma2014adam} as the optimization function with a $10^{-4}$ learning rate. The network is trained for 20 epochs, and we save the model parameters with the lowest loss in a validation set. Each batch contains 16 images along with the maps of the handcrafted signals and their ground-truth masks. We run our experiments on a Linux server with an Intel Xeon E5-2620v2 CPU, 128GB RAM, and an Nvidia GTX 1080 GPU.

\subsection{Evaluation metrics}
\label{sec:eval_metrics}
We use the pixel-level Area Under Curve (AUC) of the Receiver Operating Characteristic (ROC) as our primary metric to capture the model's performance and for comparison against the state-of-the-art splicing localization methods.

\section{Experiments and Results}

In this section, we provide an ablation study for our proposed method under various configurations (Section \ref{sec:ablation}), the comparison against state-of-the-art methods (Section \ref{sec:soa}), and some qualitative results (Section \ref{sec:qualitative}).

\subsection{Ablation study}
\label{sec:ablation}

\subsubsection{Impact of each handcrafted signal.}
First, we examine the impact of each handcrafted signal, separately and combined, fused with our Multi-Stream (MS) scheme and baseline Multi-Channel (MC) approach, where the signals are concatenated along the image channels. For the MS runs, we have two streams where one handcrafted signal is used and three streams when all inputs are combined. For the MC runs, the network has a single stream in all cases. Table \ref{tab:mc_ms_comparison} illustrates the results on the three evaluation datasets for several handcrafted signal combinations using different fusion schemes. In general, using the MS fusion scheme leads to better results than using MC for the majority of the handcrafted signals and datasets. RGB+SB with MS consistently achieves very high performance, being among the top ranks in all datasets. It outperforms its MC counterpart, achieving significantly better results on the IFS-TC dataset. Additionally, RGB+DCT+SB with MS outperforms the corresponding run with MC in all datasets, highlighting that fusing multiple signals using MS leads to better accuracy. MS-DCT reports improved performance compared to the MC-DCT on two datasets, but it is worse than the other two configurations. Additionally, combining handcrafted signals with the RGB images improves performance in general, especially in the IFS-TC dataset. Finally, DCT and SB achieve competitive performance in two datasets. This indicates that they capture useful information, which our MS architecture exploits to further improve results.

\begin{table}[t]
    \begin{center}
    \caption{Performance of our method with three signal and two fusion schemes on CASIA, IFS-TC, and Columbia. MS and MC stand for Multi-Stream and Multi-Channel processing, respectively.} \label{tab:mc_ms_comparison}
    \scalebox{0.97}{
    \begin{tabular}{|l|c|c|c|c|}
      \hline
      \textbf{Signals} & \textbf{Fus.} & \textbf{CASIA} & \textbf{IFS-TC} & \textbf{Columbia} \\ \hline\hline
      \textbf{DCT}                               & -  & 0.743 & 0.646 & 0.640 \\
      \textbf{SB}                                &-  &  0.689 & 0.750 & 0.830 \\ 
      \textbf{RGB}                               & -  & 0.877 & 0.614 & 0.818 \\ \hline\hline
      \multirow{2}{*}{\textbf{RGB+DCT}}          & MC & 0.866 & 0.732 & 0.688 \\
                                                 & MS & 0.873 & 0.689 & 0.777 \\ \hline\hline
      \multirow{2}{*}{\textbf{RGB+SB}}           & MC & 0.869 & 0.679 & 0.855 \\
                                                 & MS & 0.898 & 0.776 & 0.836 \\ \hline\hline
      \multirow{2}{*}{\textbf{RGB+DCT+SB}}       & MC & 0.851 & 0.721 & 0.717 \\
                                                 & MS & 0.873 & 0.759 & 0.782 \\ \hline
    \end{tabular}}
    \end{center}
\end{table}

\begin{table}[t]
    \begin{center}
    \caption{Performance of our method with three signals with and without fine-tuning on CASIA, IFS-TC, and Columbia. Note that we do not fine-tune our model on Columbia due to its small size.} \label{tab:fine_tuning}
    \scalebox{0.97}{
    \begin{tabular}{|l|c|c|c|c|}
      \hline
      \textbf{Signals} & \textbf{FT} & \textbf{CASIA} & \textbf{IFS-TC} & \textbf{Columbia} \\ \hline\hline
      \multirow{2}{*}{\textbf{RGB}}         & \ding{55}  & 0.765& 0.470 & 0.818  \\
                                            & \checkmark & 0.877 & 0.614  & -    \\ \hline\hline
      \multirow{2}{*}{\textbf{RGB+DCT}}     & \ding{55}  & 0.868 & 0.507 & 0.777 \\
                                            & \checkmark & 0.873 & 0.689 & -     \\ \hline\hline
      \multirow{2}{*}{\textbf{RGB+SB}}      & \ding{55}  & 0.753 & 0.460 & 0.836 \\
                                            & \checkmark & 0.898 & 0.776 & -     \\ \hline\hline
      \multirow{2}{*}{\textbf{RGB+DCT+SB}}  & \ding{55}  & 0.887 & 0.497 & 0.782 \\
                                            & \checkmark & 0.873 & 0.759 & -     \\ \hline
    \end{tabular}}
    \end{center}
\end{table}

\subsubsection{Impact of fine-tuning.}
Furthermore, we benchmark the performance of the proposed multi-stream approach with and without fine-tuning on the evaluation datasets. Table \ref{tab:fine_tuning} displays the results of our method on the three evaluation datasets when using the pre-trained and fine-tuned versions. Keep in mind that we do not fine-tune for the Columbia dataset. It is noteworthy that there is substantial performance gain in almost all cases where fine-tuning is applied. A reasonable explanation is that, with the fine-tuning on the evaluation datasets, the network learns to capture the information from the handcrafted features based on the specific domain expressed by each dataset. Therefore, the extracted cues from the employed handcrafted features might not be generalizable across different datasets. We might improve the performance further on the Columbia dataset if we could fine-tune our model on a dataset from a similar domain.

\subsubsection{Impact of skip connections.}
Additionally, we benchmark the performance of the proposed multi-stream approach with different configurations for the skip connection. Table \ref{tab:skip} displays the results of our method on the three evaluation datasets using no, image-only and all-streams skip connections. It is noteworthy that the methods perform very robustly when a skip connection is used in the image stream only. It achieves the best AUC in all cases, except for RGB+DCT+SB in the IFS-TC dataset. Finally, the experiments with no use of skip connections lead to the worst results, indicating that, thanks to the skip connections, the network learns to successfully propagate useful information from the encoder streams to the decoder. Yet, skip connections from the handcrafted signals do not always help. 

\begin{table}[t]
    \begin{center}
    \caption{Performance of our method with three signals and three configurations for the skip connection on CASIA, IFS-TC, and Columbia. \textit{No} indicates that no skip connections are used. \textit{Img} indicates that skip connection is used only for the image stream. \textit{All} indicates that skip connections are used only for all streams.} 
    \label{tab:skip}
    \scalebox{0.97}{
    \begin{tabular}{|l|c|c|c|c|}
      \hline
      \textbf{Signals} & \textbf{Skip} & \textbf{CASIA} & \textbf{IFS-TC} & \textbf{Columbia} \\ \hline\hline
      \multirow{3}{*}{\textbf{RGB+DCT}}     & \textbf{\textit{No}}    & 0.857 & 0.732 & 0.623 \\
                                            & \textbf{\textit{Img}}   & 0.873 & 0.689 & 0.777 \\
                                            & \textbf{\textit{All}}   & 0.840 & 0.616 & 0.742 \\ \hline\hline
      \multirow{3}{*}{\textbf{RGB+SB}}      & \textbf{\textit{No}}    & 0.879 & 0.773 & 0.741 \\
                                            & \textbf{\textit{Img}}   & 0.898 & 0.776 & 0.836 \\
                                            & \textbf{\textit{All}}   & 0.882 & 0.718 & 0.826 \\ \hline\hline
      \multirow{3}{*}{\textbf{RGB+DCT+SB}}  & \textbf{\textit{No}}    & 0.797 & 0.763 & 0.566 \\
                                            & \textbf{\textit{Img}}   & 0.873 & 0.759 & 0.782 \\
                                            & \textbf{\textit{All}}   & 0.871 & 0.808 & 0.762 \\ \hline
    \end{tabular}}
    \end{center}
\end{table}

\subsection{Comparison with the state-of-the-art}
\label{sec:soa}

In Table \ref{tab:compare}, we present our evaluation in comparison to four state-of-the-art approaches. We select our networks with MS fusion and with skip connection only to the image stream, denoted as MS-DCT, MS-SB, and MS-DCT+SB for the three signal combinations. As state-of-the-art approaches, we have re-implemented three methods, LSTMEnDec \cite{Bappy2019}, LSTMEnDecSkip \cite{Mazaheri2019}, and OwAF \cite{Charitidis2021}, using the same training pipeline as the one for the development of our networks for fair comparison. These are closely related methods to the proposed one. Also, we benchmark against the publicly available PyTorch implementation of ManTra-Net \cite{Wu2019}\footnote{\url{https://github.com/RonyAbecidan/ManTraNet-pytorch}} without fine-tuning it on the evaluation datasets. All methods are benchmarked on the same evaluation sets. In general, all three variants of our method achieve competitive performance on all evaluation datasets, outperforming the state-of-the-art approaches in several cases with a significant margin. Our MS-SB leads to the best results with 0.898 AUC, respectively, with the second-best LSTMEnDecSkip approach achieving 0.810. Similar results are reported on the IFS-TC dataset. Our MS-SB achieves the best AUC with 0.776, followed by the OwAF method with 0.680. Finally, our MS-SB achieves the best results in the Columbia dataset with 0.836. Notably, the LSTMEnDec is the second-best approach, outperforming our two other variants, MS-DCT and MS-DCT+SB; however, this method performs poorly on the other two datasets.

\begin{table}[t]
\begin{center}
\caption{Performance comparison against the state-of-the-art on CASIA, IFS-TC, and Columbia.} \label{tab:compare}
    \scalebox{0.97}{
    \begin{tabular}{|l|c|c|c|}
      \hline
      \textbf{Method} & \textbf{CASIA} & \textbf{IFS-TC} & \textbf{Columbia} \\ \hline\hline
      \textbf{ManTra-Net} \cite{Wu2019}          & 0.665 & 0.547 & 0.660 \\
      \textbf{LSTMEnDec} \cite{Bappy2019}        & 0.628 & 0.648 & 0.809 \\
      \textbf{LSTMEnDecSkip} \cite{Mazaheri2019} & 0.810 & 0.670 & 0.207 \\
      \textbf{OwAF} \cite{Charitidis2021}        & 0.754 & 0.680 & 0.551 \\ \hline
      \textbf{MS-DCT} (Ours)                     & 0.873 & 0.689 & 0.777 \\
      \textbf{MS-SB} (Ours)                      & \textbf{0.898} & \textbf{0.776} & \textbf{0.836} \\
      \textbf{MS-DCT+SB} (Ours)                  & 0.873 & 0.759 & 0.782 \\ \hline
    \end{tabular}}
\end{center}
\end{table}

\subsection{Qualitative Results}
\label{sec:qualitative}

\begin{figure*}
    \centering
    \includegraphics[width=\linewidth]{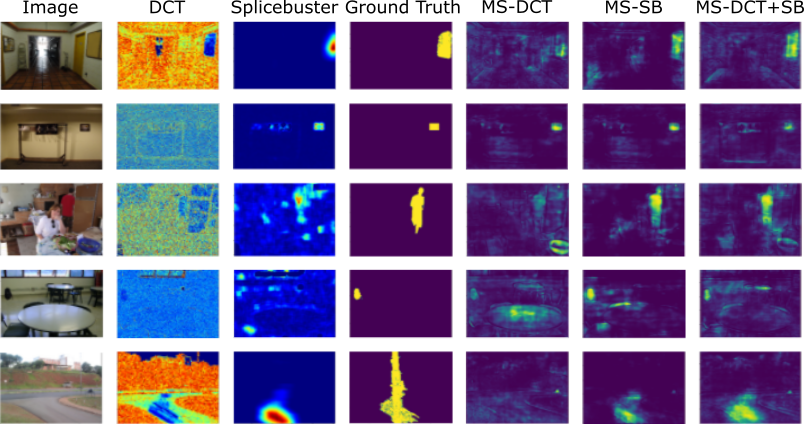}
    \caption{Visual examples of our multi-stream network with three signal combinations from the IFS-TC dataset.}
    \label{fig:qifstc}
\end{figure*}

Figure \ref{fig:qifstc} illustrates some example results from the IFS-TC dataset. The first three columns contain the network inputs, i.e., the RGB image, DCT, and Splicebuster. The third column presents the ground truth masks, and the last ones depict the network predictions. In the first example, Splicebuster provides a useful lead to the network, which is able to detect the tampered area with high accuracy, especially the MS-SB run. In the second case, all of our networks detect the tampered area, even though DCT does not seem to be helpful. 
In the next two examples, none of the handcrafted signals precisely localize splicing, but our MS-SB and MS-DCT+SB are able to detect it partially. Finally, in the last case, our networks failed to localize the forged areas in the image, although the two handcrafted signals highlight the correct area only in a small part. In general, the qualitative results here align with the quantitative of the previous sections, with MS-DCT providing the worst predictions among our three settings, while MS-SB detects the tampered areas with significantly higher accuracy.

\section{Conclusion}
\label{sec:concl} 

In this work, we proposed a deep learning method that localizes spliced regions in images by fusing features extracted from the RGB images with ones extracted from handcrafted signals based on a multi-stream fusion pipeline. We experimented with two popular handcrafted signals based on DCT and Splicebuster algorithms. Through an ablation study on three datasets, we demonstrated that our multi-stream fusion approach yields competitive performance consistently. Also, we compared our approach to four state-of-the-art methods, achieving the best performance on all three datasets. In the future, we plan to investigate more architectural choices that improve the effectiveness of signal fusion and employ more robust handcrafted signals.

\bigskip
\noindent\textbf{Acknowledgments}:
This research has been supported by the H2020 MediaVerse and Horizon Europe vera.ai projects, which are funded by the European Union under contract numbers 957252 and 101070093.

%
%
%
\bibliographystyle{splncs04}
\bibliography{bibliography}
\end{document}